\documentclass[conference]{IEEEtran}
\IEEEoverridecommandlockouts
\usepackage{cite}
\usepackage{hyperref}
\usepackage{amsmath,amssymb,amsfonts}
\usepackage{graphicx}
\usepackage{textcomp}
\usepackage{algorithm}
\usepackage{tabularx}

\usepackage{algorithmic}
\usepackage{amsmath}
\usepackage{xcolor}
\usepackage{booktabs}                  %表格线条加粗所需
\def\BibTeX{{\rm B\kern-.05em{\sc i\kern-.025em b}\kern-.08em
    T\kern-.1667em\lower.7ex\hbox{E}\kern-.125emX}}
\begin{document}

\title{CADE: Cosine Annealing Differential Evolution for Spiking Neural Network\\
% {\footnotesize \textsuperscript{*}Note: Sub-titles are not captured in Xplore and
% should not be used}
% \thanks{Identify applicable funding agency here. If none, delete this.}
}

\author{
\IEEEauthorblockN{
Runhua Jiang\IEEEauthorrefmark{2}*,
Guodong Du\IEEEauthorrefmark{3}*, 
Shuyang Yu\IEEEauthorrefmark{2}, 
Yifei Guo\IEEEauthorrefmark{2},
Sim Kuan Goh\IEEEauthorrefmark{2}, 
Ho-Kin Tang\IEEEauthorrefmark{3}
}

\IEEEauthorblockA{\textit{\IEEEauthorrefmark{2}School of Electrical Engineering and Artificial Intelligence, Xiamen University Malaysia}, Selangor, Malaysia}

% Email: \{ait2009366, ait2109105, ait2109083, simkuan.goh\}@xmu.edu.my}
\IEEEauthorblockA{\textit{\IEEEauthorrefmark{3}School of Science, Harbin Institute of Technology (Shenzhen),}
Shenzhen, China.\\ }
*contributed equally to this paper.

% Email: \{duguodong7@gmail.com, denghaojian@hit.edu.cn\}}

\thanks{This research is supported by the Ministry of Higher Education Malaysia through the Fundamental Research Grant Scheme (FRGS/1/2023/ICT02/XMU/02/1), and Xiamen University Malaysia through Xiamen University Malaysia Research Fund (XMUMRF/2022-C10/IECE/0039 and XMUMRF/2024-C13/IECE/0049). This work also acknowledges support from the National Natural Science Foundation of China~(Grant No.  12204130), Shenzhen Start-Up Research Funds~(Grant No. HA11409065) and the HITSZ Start-Up Funds~(Grant No. X2022000). Corresponding: simkuan.goh@xmu.edu.my, denghaojian@hit.edu.cn}
}

\maketitle
% \footnotetext[1]{Source Code:\url{https://github.com/Tank-Jiang/Enhance-Spiking-Wise-Element-by-using-Cosine-Adaptive-Differential-Evolution}.}

\begin{abstract}

Spiking neural networks (SNNs) have gained prominence for their potential in neuromorphic computing and energy-efficient artificial intelligence, yet optimizing them remains a formidable challenge for gradient-based methods due to their discrete, spike-based computation. This paper attempts to tackle the challenges by introducing Cosine Annealing Differential Evolution (CADE), designed to modulate the mutation factor (F) and crossover rate (CR) of differential evolution (DE) for the SNN model, i.e., Spiking Element Wise (SEW) ResNet. Extensive empirical evaluations were conducted to analyze CADE. CADE showed a balance in exploring and exploiting the search space, resulting in accelerated convergence and improved accuracy compared to existing gradient-based and DE-based methods.
Moreover, an initialization method based on a transfer learning setting was developed, pretraining on a source dataset (i.e., CIFAR-10) and fine-tuning the target dataset (i.e., CIFAR-100), to improve population diversity. It was found to further enhance CADE for SNN. Remarkably, CADE elevates the performance of the highest accuracy SEW model by an additional 0.52 percentage points, underscoring its effectiveness in fine-tuning and enhancing SNNs. These findings emphasize the pivotal role of a scheduler for F and CR adjustment, especially for DE-based SNN. Source Code on Github: \url{https://github.com/Tank-Jiang/CADE4SNN}.

\end{abstract}

\section{Introduction}

In the last decade, there has been a huge development in artificial neural network theories and applications \cite{Lee2021Energy-efficient,ann1,ann2}. It begins with the first-generation multi-layer perceptron to the many state-of-the-art techniques in the second-generation deep neural networks (DNNs) trained using gradient descent. Despite this great advancement, ANNs still lag behind the biological neural networks in terms of energy efficiency and abilities for online learning. Though ANNs are brain-inspired, they are fundamentally different in structure, neural computations, and learning rules compared to the biological neural network. This leads to the third generation of neural networks, spiking neural networks (SNNs), which are considered a breakthrough in addressing the bottlenecks of ANNs. Unlike ANNs that continuously compute over all nodes, SNNs only activate specific neurons in response to stimuli, significantly reducing energy consumption \cite{Lee2021Energy-efficient}. This attribute is particularly beneficial for processing sensor data or performing tasks where only a subset of the data changes over time. Moreover, SNNs excel in dynamic and evolving environments through mechanisms such as online learning, enabling them to adapt continuously and in real-time to new data. This makes SNNs highly effective for applications that require immediate processing of incoming data streams without the need to pause for retraining, as they leverage updated models instantaneously \cite{Xiao2022Online}. With the ability to address the major challenges of deep neural networks, such as high resource requirements including energy consumption, data storage, and computational costs, Spiking Neural Networks (SNNs) \cite{christensen20222022,kundu2021hire} have emerged as a promising alternative to conventional Artificial Neural Networks (ANNs) in recent years.

% In the last decade, there has been a huge development in artificial neural network architecture. It begins with the first-generation multi-layer perceptron to the many state-of-the-art techniques in the second-generation deep neural networks (DNNs) trained using gradient descent. Despite this great advancement, ANNs still lag behind the biological neural networks in terms of energy efficiency and abilities for online learning. Though ANNs are brain-inspired, they are fundamentally different in structure, neural computations, and learning rules compared to the biological neural network. This leads to the third generation of neural networks, spiking neural networks (SNNs). SNNs are considered as the breakthrough of bottlenecks of ANNs. Though there has been significant success in the field of deep neural networks across various domains. However, one of the major challenges associated with deep neural networks is their high resource requirements, including energy consumption, data storage, and computational costs. In recent years, Spiking Neural Networks (SNNs) (Jaiswal \& Panda, 2019; Christensen et al., 2022; Wu et al.,2018; Kundu et al., 2021; Fang et al., 2021; Roy et al., 2019) have emerged as a promising alternative to conventional Artificial Neural Networks (ANNs) due to their potential for energy efficiency.

However, the training of SNNs can be challenging due to their non-differentiable nature of spiking events. Most prior SNN methods use ANN-like architectures (e.g., VGGNet or ResNet), which could provide sub-optimal performance for temporal sequence processing of binary information in SNNs. Traditional gradient-based learning methods often suffer when it comes to training SNN. An alternative to gradient-based methods Evolution algorithms (EAs). By using EAs to optimize the weights and architecture of SNNs, researchers can bypass the need for gradient computation, offering a robust alternative to train SNNs for a variety of tasks. The combination of DE's global search capability and SNN's efficient information processing holds promise for developing advanced computational models that are both powerful and energy-efficient.

EAs are grounded in Darwinian evolutionary theory and are recognized for their effectiveness in tackling complex problems. They mimic the natural process of survival of the fittest within a population, displaying remarkable robustness and adaptability in identifying global solutions to various optimization challenges. The initial development of evolutionary computing traces back to the 1950s with the introduction of the Genetic Algorithm (GA) \cite{holland75}, which incorporates Charles Darwin's theory of natural selection. Subsequent decades, particularly the 1960s and 1970s, saw the emergence of other evolutionary algorithm forms like Evolutionary Programming (EP) \cite{fogel66} and Evolution Strategies (ES) \cite{hansen15}. While Differential Evolution (DE) has been applied in optimizing Deep Neural Networks (DNNs), the effectiveness of this approach in training conventional DNN models is not fully established. This is partly because most existing research tends to focus on custom-designed network architectures rather than standard ones.

In this work, we hypothesize existing SEW that compete for survival and breeding in evolution will show higher accuracy and stronger tolerance to data set and design an experiment, to examine the influence of EA on DNNs’ corruption robustness. Specifically, pre-trained SEW CIFAR-100 evolved without changing the original dataset, loss function, and network architecture.

\section{Related Works}
\begin{figure*}[ht]
\centering
\includegraphics[width=\textwidth]{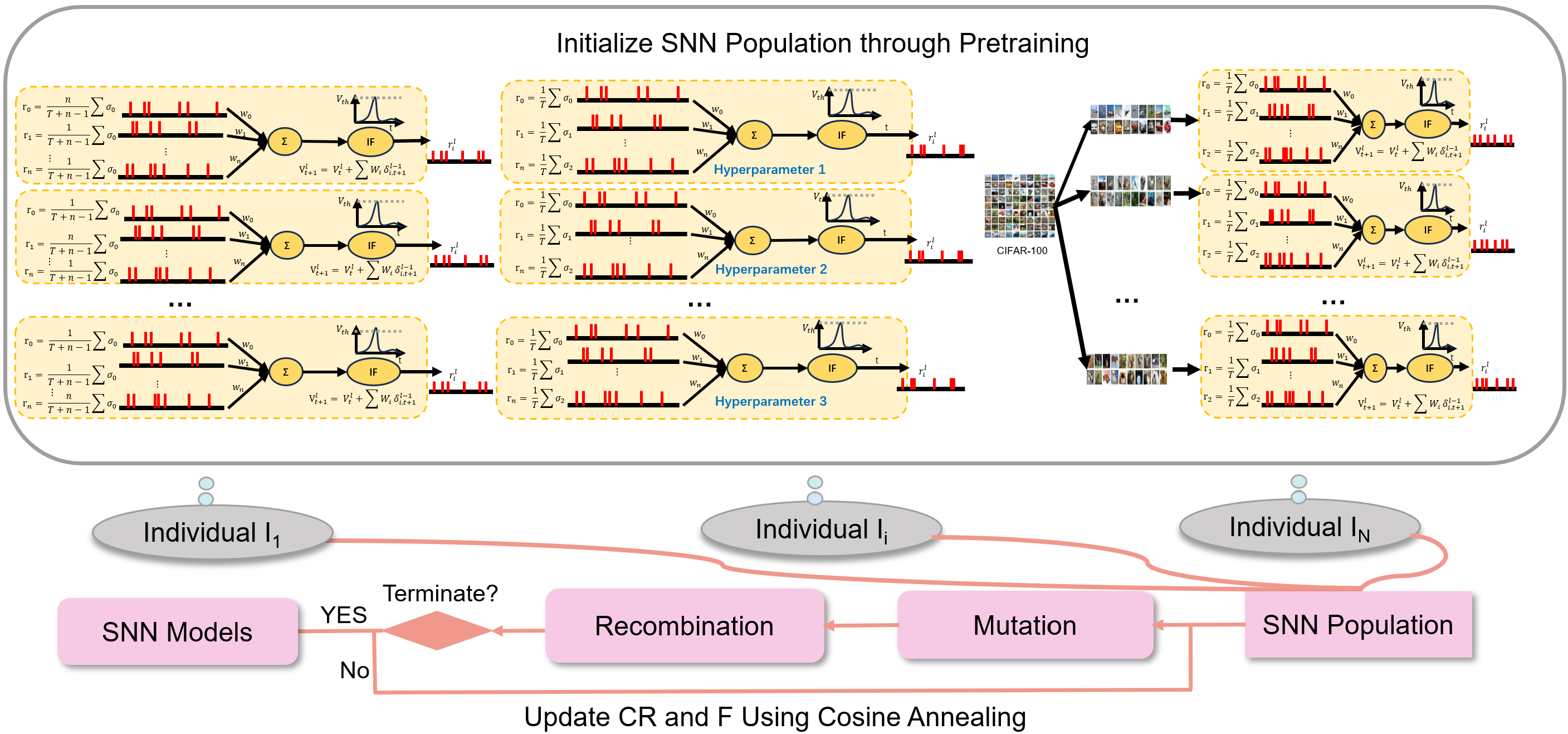}
\caption{Pipeline of the proposed CADE for Spiking Neural Network.}
\label{fig:concept_digram}
\end{figure*}

\subsection{Training methods for Spiking Neural Network}
Recent research revealed that deep spiking neural networks (SNNs) have the potential for improving the latency and energy efficiency of deep neural networks through data-driven event-based computation\cite{ji2023sctn}.
However, SNNs have not quite reached the same accuracy levels of ANNs in traditional machine learning tasks practically. A main reason for this is the lack of efficient training algorithms for deep SNNs due to the non-differentiable architecture. Meanwhile current adequate algorithms are well prepared for deep neural networks. There is a research gap here. Spike-Time-Dependent Plasticity (STDP) \cite{liu21} is a learning rule commonly used in spiking neural networks (SNNs) to update the synaptic weights between neurons based on the precise timing of pre- and postsynaptic spikes. It is inspired by the observed phenomenon in biological neurons where the strength of synaptic connections is modulated by the relative timing of pre- and post-synaptic spikes. 

To deal with this non-differentiable transfer function, various methods have been proposed to address the non-differentiable transfer function. On the one side, some researches improve the standard backpropagation. On the other side, non-differentiable spike generation function was handled using surrogate models (e.g., Gaussian process~\cite{gp0,gp1,gp2}). A novel method enables error backpropagation in deep spiking neural networks (SNNs) by treating neuron membrane potentials as differentiable signals and spike discontinuities as noise\cite{wunderlich2020eventprop}. This approach, which uses principles from optimal control theory, supports efficient training across various SNN architectures without needing approximations. % 这段话修改了，换了新的cite
Spike-based backpropagation \cite{lee20} is proposed that can enable training deep convolutional SNNs directly (with input spike events). This derivative method can work well for leaky behavior of Leaky integrate and fire (LIF) neurons.

% One the other side, recently the indirect methods are proposed that is to use different data representations between training and processing, i.e., training a conventional ANN and developing conversion algorithms that transfer the weights into equivalent deep SNNs. Such as Siegert approximation for Integrate-and-Fire neurons to map an offline-trained DBN onto an efficient event-driven spiking neural network \cite{oconnor13}. 
On the other side, recently proposed indirect methods involve using different data representations between training and processing. For example, training a conventional artificial neural network (ANN) and then using conversion algorithms to transfer the weights into equivalent deep spiking neural networks (SNNs). This includes advanced techniques such as Logarithmic Temporal Coding, which efficiently reduces computational costs by adjusting the spike encoding during the conversion of a trained ANN to an SNN, offering a streamlined method for achieving competitive classification performance with reduced computational demand\cite{zhang2020efficient}.
But converting deep ANNs into SNNs may have come at the cost of performance losses without a notion of time, to sparsely firing, event-driven SNNs. 
% To deal with those problems, a set of optimization techniques to minimize performance loss in the conversion process for ConvNets and fully connected deep networks \cite{diehl15}. 
To address these issues, recent advancements introduce optimization methods that minimize performance loss during the conversion of ConvNets and fully connected networks into SNNs, significantly improving both accuracy and latency\cite{wang2022}.
However, the main problem is that details of statistics in spike trains that go beyond ideal mean rate modeling, such as required for processing practical event-based sensor data cannot be precisely represented by the signals used for training. So far it has only been possible to train single layers with the learning rules operating directly on spike trains. 

\subsection{Differential Evolution Algorithm}
Differential Evolution is a robust, stochastic evolutionary algorithm that has gained widespread recognition for its efficacy in solving complex optimization problems across diverse domains. However, it faces a significant challenge in the realm of hyperparameter tuning. Hyperparameters in DE, primarily comprising the population size, mutation factor, and crossover rate, play a crucial role in dictating the algorithm's performance. 

Consequently, numerous research efforts have been initiated to propose various mechanisms to address the hyperparameter challenge in DE. For instance, the SADE algorithm includes an adaptive mechanism that autonomously adjusts strategy parameters like mutation and crossover strategies and rates during the evolutionary process \cite{omran05}. This self-adaptive approach allows SADE to dynamically tailor its behavior according to the specific requirements of the optimization problem, thereby enhancing its efficiency and robustness. Similarly, the Success-History based Adaptive Differential Evolution (SHADE) algorithm introduces a sophisticated adaptation technique that relies on historical performance data \cite{tanabe13}. SHADE maintains a memory that records successful parameter settings from previous generations. These records are then used to guide the mutation and crossover rates, enabling a more informed and targeted search process. By incorporating this historical success information, SHADE effectively navigates complex optimization landscapes, achieving superior convergence rates and solution quality compared to traditional DE and other variants.

%加了一段话
While SADE and SHADE were innovative, the field has evolved with new algorithms that integrate machine learning to enhance performance in complex scenarios, often outperforming traditional methods \cite{morales2019differential}. This highlights the continuous advancement and dynamic nature of optimization technologies.

\subsection{Robustness}
In the field of computer vision, a key objective is to enhance deep learning systems to rival or surpass the robustness of the human visual system. This pursuit involves addressing various challenges, with researchers exploring different approaches. Carilini and Wagner have made strides in increasing the resilience of networks against adversarial examples, while Hendrycks et al. have concentrated on dealing with unknown unknowns, anomalous inputs, and bolstering networks against input corruption [14]. In this context, the introduction of datasets like Imagenet-C, featuring 75 common visual corruptions at varying intensities, and CIFAR-100-C, designed for assessing corruption robustness, has been significant.

Moreover, the exploration of spiking neural networks (SNNs) has opened new avenues in the quest for robustness. SNNs, which more closely mimic the biological processes of the human brain, offer potential advantages in terms of energy efficiency and latency. Their inherent temporal dynamics and spike-based information processing can lead to enhanced robustness against certain types of noise and distortions, making them a promising area of research in developing resilient deep learning models for computer vision tasks.

\section{Proposed Work}
In our work, we propose the Cosine Annealing Differential Evolution (CADE) algorithm to improve the accuracy and robustness of the SEW model without changing the original dataset, loss function, and network architecture, illustrated in Fig.~\ref{fig:concept_digram}.
Subsequently, the overall performance of robustness of our proposed optimization scheme is evaluated on CIFAR-100-C.
\subsection{CADE Algorithm}
DE is a globally-oriented stochastic optimization technique that operates on a population basis. Its primary function is to conduct an extensive search for optimal solutions. In DE, a set of potential solutions, referred to as "individuals," constitutes the population. Each individual represents a candidate solution to the optimization problem. The evolution of these individuals is a systematic process, where new candidate solutions emerge through a "mutation" process. This is subsequently followed by a "crossover" phase, which amalgamates these new solutions with the existing individuals, thus forming a fitter population. This iterative cycle of evolution continues until the process either reaches a pre-set number of iterations or the optimization goal attains a level of convergence deemed satisfactory.

The mutation factor (F), and crossover rate (CR) are critical hyperparameters in the DE algorithm. Properly tuning these hyperparameters is essential to strike a balance between exploration and exploitation, leading to efficient and effective optimization for various real-world problems. However, the optimal setting of these parameters is not straightforward and is highly problem-dependent, leading to the primary challenge in DE applications. 
Regarding the hyperparameter issue, this paper also proposes a variant of the DE algorithm. In CADE, the F and CR are dynamically varied throughout the evolutionary process, following the pattern of a cosine function. This concept is inspired by the annealing learning rate approach, where the learning rate is adjusted using a cosine function. The algorithm pseudocode is shown in the algorithm \ref{enhanced_de_algorithm}.

\begin{algorithm}
\caption{CADE Algorithm}
\label{enhanced_de_algorithm}
\begin{algorithmic}
\STATE Initialize population $P$ with random solutions
\STATE Initialize parameters: $F_{\text{min}}, F_{\text{max}} = F_{\text{init}}, CR_{\text{min}}, CR_{\text{max}} = CR_{\text{init}} $
\STATE Set termination criteria: max iterations, convergence threshold, etc.
\STATE Initialize current iteration $t = 0$

\WHILE{not reached termination criteria}
    \FOR{each individual $x_i$ in population $P$}
        \STATE Randomly select three distinct individuals $a, b, c$ from $P$
        \STATE Generate a trial solution $u$ using the mutation strategy: $u = a + F \cdot (b - c)$
        \STATE Apply crossover with probability \\ $CR$: $v_j = \begin{cases} u_j, & \text{if } \text{rand()} \leq CR \text{ or } j = \text{rand()}(1, D) \\ x_{ij}, & \text{otherwise} \end{cases}$
        \STATE Evaluate the fitness of $v$
        \IF{$v$ is better than $x_i$}
            \STATE Replace $x_i$ with $v$ in population $P$
        \ENDIF
    \ENDFOR
    \STATE Update iteration counter: $t = t + 1$\\
    \textbf{Strategy 1, 2, 3 or 4} for CR and F update
    % \STATE $F = F_{\text{min}} + \frac{F_{\text{max}} - F_{\text{min}}}{2} \left(1 + \cos\left(\frac{\pi \cdot t}{\text{max\_iterations}}\right)\right)$
    % \STATE $CR = CR_{\text{min}} + \frac{CR_{\text{max}} - CR_{\text{min}}}{2} \left(1 + \cos\left(\frac{\pi \cdot t}{\text{max\_iterations}}\right)\right)$
\ENDWHILE

\RETURN Best solution found in $P$
\end{algorithmic}
\end{algorithm}
\subsection{Strategies for CR and F update}
Hyperparameter tuning plays a pivotal role in the performance of differential evolution algorithms, impacting their optimization capabilities across various iterations. Traditional approaches often rely on static or manually adjusted hyperparameters, which may not optimally adapt to dynamic problem landscapes encountered during optimization processes. This paper introduces a novel update strategy that utilizes a cosine function transformation to dynamically adjust hyperparameters during iterations. By implementing this strategy, the algorithm can search and adapt to the most suitable hyperparameters, thereby enhancing model optimization effectively. We provide a comprehensive analysis of how different hyperparameters influence the algorithm's performance and demonstrate the superiority of our proposed cosine-based adaptation strategy through extensive experiments on benchmark functions. The results indicate significant improvements in convergence rates and solution accuracy, underscoring the importance of adaptive hyperparameter strategies in evolutionary computations.

To explore further the update strategy, this work consider four distinct approaches to dynamically adjust the parameters CR and F within the Differential Evolution algorithm framework. Each strategy offers a different method for parameter adaptation, potentially impacting the algorithm's ability to navigate the search space and converge on optimal solutions.

\textbf{Strategy 1} updates both CR and F according to a cosine function that depends on the current iteration (t) and the maximum number of iterations (max\_iterations), if the new individual (a solution candidate) does not lead to a better result. The updates aim to adjust the parameters dynamically over the course of the algorithm's run, enabling the balance between exploration and exploitation.
\floatname{algorithm}{Strategy} % Changes the name from 'Algorithm' to 'Strategy'

\setcounter{algorithm}{0}

\begin{algorithm}
\caption{Update rules for CR and F}
\label{Strategy1}
\begin{algorithmic}
\IF{New individual is not better} 
    \STATE \( F = F_{\text{min}} + \frac{F_{\text{max}} - F_{\text{min}}}{2} \left(1 + \cos\left(\frac{\pi \cdot t}{\text{max\_iterations}}\right)\right) \)
    \STATE  \( CR = CR_{\text{min}} + \frac{CR_{\text{max}} - CR_{\text{min}}}{2} \left(1 + \cos\left(\frac{\pi \cdot t}{\text{max\_iterations}}\right)\right) \)
\ENDIF
\end{algorithmic}
\end{algorithm}
\textbf{Strategy 2} straightforwardly updates CR and F using the same cosine function as in Strategy 1, but it does not condition the update on the performance of the new individual. This suggests a consistent, periodic adjustment of the parameters regardless of the immediate outcomes.
\begin{algorithm}
\caption{Update rules for CR and F}
\label{Strategy2}
\begin{algorithmic}
\IF{New individual is not better} 
    \STATE \( F = F_{\text{min}} + \frac{F_{\text{max}} - F_{\text{min}}}{2} \left(1 + \cos\left(\frac{\pi \cdot t}{\text{max\_iterations}}\right)\right) \)
    \STATE  \( CR = CR_{\text{min}} + \frac{CR_{\text{max}} - CR_{\text{min}}}{2} \left(1 + \cos\left(\frac{\pi \cdot t}{\text{max\_iterations}}\right)\right) \)
\ENDIF
\end{algorithmic}
\end{algorithm}

CR and F are updated as in Strategy 2, but after the deterministic update, a stochastic component is added by using random value in \textbf{strategy 3}, which introduces randomness into the new values of CR and F. The random.uniform() function generates a random number between the given min and max values for each parameter, adding an element of randomness to the updating process.
\begin{algorithm}
\caption{Update rules for CR and F}
\begin{algorithmic}
 
\STATE \( F = F_{\text{min}} + \frac{F_{\text{max}} - F_{\text{min}}}{2} \left(1 + \cos\left(\frac{\pi \cdot t}{\text{max\_iterations}}\right)\right) \)
\STATE \( F = F + \text{random.uniform}(F_{\text{min}}, F_{\text{max}}) \)
\STATE \( CR = CR_{\text{min}} + \frac{CR_{\text{max}} - CR_{\text{min}}}{2} \left(1 + \cos\left(\frac{\pi \cdot t}{\text{max\_iterations}}\right)\right) \)
\STATE \( CR = CR + \text{random.uniform}(CR_{\text{min}}, CR_{\text{max}}) \)
\end{algorithmic}
\end{algorithm}
\textbf{Strategy 4} combines the conditional approach of Strategy 1 with the stochastic element of Strategy 3. If the new individual is not better, both CR and F are first updated deterministically using the cosine function, and then they are further adjusted by adding a random value from a uniform distribution. This strategy seems to adapt the parameters based on the current success of the algorithm and introduces variability to escape local optima or to search new areas of the solution space.
\begin{algorithm}
\caption{Update rules for CR and F}
\begin{algorithmic}
\IF {New individual is not better}
    \STATE \( F = F_{\text{min}} + \frac{F_{\text{max}} - F_{\text{min}}}{2} \left(1 + \cos\left(\frac{\pi \cdot t}{\text{max\_iterations}}\right)\right) \)
    \STATE \( F = F + \text{random.uniform}(F_{\text{min}}, F_{\text{max}}) \)
    \STATE \( CR = CR_{\text{min}} + \frac{CR_{\text{max}} - CR_{\text{min}}}{2} \left(1 + \cos\left(\frac{\pi \cdot t}{\text{max\_iterations}}\right)\right) \)
    \STATE \( CR = CR + \text{random.uniform}(CR_{\text{min}}, CR_{\text{max}}) \)
\ENDIF
\end{algorithmic}
\end{algorithm}

\subsection{Spiking Element Wise Model}
Deep Spiking Neural Networks (SNNs) pose unique challenges for optimization using gradient-based methods due to their discrete, binary activations and intricate spatial-temporal dynamics. Given the remarkable achievements of ResNet in the field of deep learning, applying residual learning to train deep SNNs emerges as a logical progression. Inspired by this, the Spike-Element-Wise (SEW) residual block have been developed \cite{fang2022deep}, specifically designed for SNNs to enable efficient residual learning. This block not only facilitates identity mapping but also effectively addresses the issues of vanishing and exploding gradients. As demonstrated in Fig. \ref{fig:sew}, the SEW residual block is illustrated as follows: 
\begin{figure}[htbp]
\centerline{\includegraphics[width=0.48\textwidth]{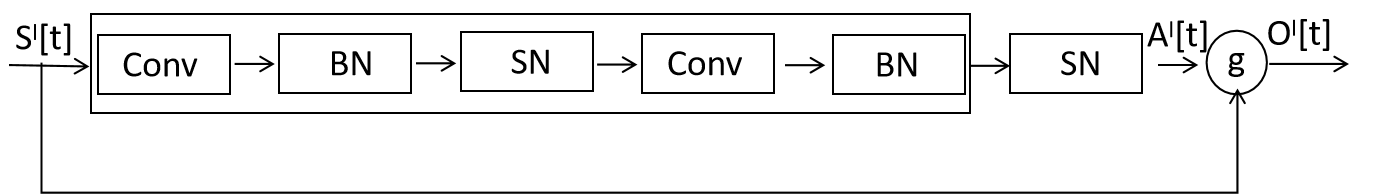}}
\caption{SEW residual block}
\label{fig:sew}
\end{figure}
\begin{equation}
O'[t] = g(\text{SN}(\text{SF}(s'[t])), s[t]), \quad S'[t] = g(A'[t], S'[t]),
\end{equation}
It's intriguing to note that SEW ResNet can easily implement identity mapping, leveraging the binary property of spikes. By capitalizing on the unique characteristics of spiking neurons and their binary spiking behavior, different element-wise functions, denoted as g, can be identified that satisfy identity mapping. The binary property of spikes introduces a distinctive approach to realizing identity mapping in SEW ResNet, offering a novel perspective on the implementation of this fundamental concept in the context of spiking neural networks.The specifics of these element-wise functions, as indicated in Table \ref{tab:element_wise_functions}.
\begin{table}[h]
\centering
\caption{List of element-wise functions \( g \).}
\label{tab:element_wise_functions}
\begin{tabular}{ll}
\hline
Name & Expression of \( g(A'[t], S[t]) \) \\
\hline
ADD & \( A'[t] + S[t] \) \\
AND & \( A'[t] \land S[t] = A'[t] \cdot S[t] \) \\
IAND & \( \neg A'[t] \land S[t] = (1 - A'[t]) \cdot S[t] \) \\
\hline
\end{tabular}
\end{table}
The two keys are formulation of downsample block and SEW ResNet can overcome vanishing/exploding gradient. For the downsample, in cases where the dimensions of the input and output of a residual block differ, the shortcut connection is adapted to perform downsampling. This is achieved by using convolutional layers with a stride greater than 1. This strategy ensures that the shortcut connection aligns with the spatial dimensions of the feature maps using spiking neurons(SN) in shortcut.
% \begin{figure}[htbp]
%   \centering
%   \begin{minipage}{0.24\textwidth}
%     \includegraphics[width=\linewidth]{graph/Downsample Basic Block.png} % 第一张图的路径
%     \caption{Downsample Basic Block}
%     \label{fig:Downsample Basic Block}
%   \end{minipage}
%   \hfill % 添加一些水平间距
%   \begin{minipage}{0.24\textwidth}
%     \includegraphics[width=\linewidth]{graph/Downsample SEW Block.png} % 第二张图的路径
%     \caption{Downsample SEW Block}
%     \label{fig:Downsample SEW Block}
%   \end{minipage}
% \end{figure}
The experimental results suggest that the ReLU
before addition (RBA) block does not perform as well as the basic block . The SEW block is described as an extension of the RBA block. This implies that the SEW block incorporates certain characteristics of the RBA block while potentially introducing additional features or modifications that address the limitations observed in the RBA block. The SEW block utilizes AND, IAND, and ADD as potential element-wise functions (g). When AND and IAND are used, the output is spikes represented as binary tensors. This ensures that the "infinite outputs problem" in Artificial Neural Networks (ANNs), where unbounded activations can lead to numerical instability, does not occur in SNNs with SEW blocks. In the experiment, SEW ResNet-34 was used.
\section{experiment}
\subsection{Dataset}
In the experimental section of our study, this work focuses on two benchmark datasets, \textbf{CIFAR-10} and \textbf{CIFAR-100}, which are widely used in the field of machine learning for evaluating image classification models. 
% The CIFAR-10 dataset is a collection of 60,000 32x32 color images, divided into 10 classes, with each class representing different objects such as cats, dogs, birds, and airplanes.

\textbf{CIFAR-100-C} stands as an extension to the CIFAR-100 dataset, designed specifically for the evaluation of model robustness. Unlike the standard CIFAR-100 dataset, CIFAR-100-C introduces a challenging set of conditions by intentionally corrupting the images with various types and levels of distortions. This dataset serves as a rigorous benchmark to assess the ability of machine learning models to generalize and perform well under diverse and adverse circumstances. 
CIFAR-100-C encompasses a broad spectrum of corruption types, including but not limited to blur, noise, and brightness variations. Each corruption type introduces a unique challenge, testing the model's resilience against a range of real-world distortions. The dataset provides images corrupted at different severity levels, allowing for a granular evaluation of a model's performance under varying degrees of perturbation. 

% This variability ensures a thorough examination of robustness across a spectrum of conditions. The intentional introduction of corruptions in CIFAR-100-C aligns with real-world scenarios where data imperfections are prevalent. 

% Evaluating models on this dataset provides insights into their capacity to handle the complexities and uncertainties associated with practical applications.
\subsection{Initial Population}
In this paper, transfer learning is used to initialize population. 
% Transfer learning is a powerful approach in the field of machine learning that involves transferring knowledge from one domain data set to another data set. 
% This technique is particularly prevalent in the realm of deep learning, where it has enabled remarkable progress in areas such as computer vision and natural language processing. In a later experiment, it employed a unique approach for initializing the population in the Differential Evolution algorithm, focusing on finetuning neural network models. 
This approach involved using pre-trained models from the CIFAR-10 dataset and subsequently fine-tuning them on the CIFAR-100 dataset. Based on the result, the transfer learning not only markedly accelerated the training process, but also improved accuracy compared to those trained from scratch or initialized differently.

% Finetuning is a specific strategy within transfer learning. It involves taking a pre-trained model (usually on a large dataset and a related task) and adapting it to the new specific requirements of a new task. Fine-tuning typically entails adjusting the weights of the pre-trained model's layers, often by adding additional layers at the end to adapt to the target task. These additional layers are usually randomly initialized and trained using the target task's data. Last but not least, Fine-tuning aims to preserve the valuable knowledge acquired during pre-training while allowing the model to adapt to the nuances of the target task.

Finetuning not only offers the advantage of time efficiency in the initial population creation but also presents a unique opportunity for diversification. The algorithm can generate individuals with distinct characteristics by employing various hyperparameter configurations during the fine-tuning phase. This diversification strategy plays a crucial role in enriching the population's genetic pool, promoting a broader exploration of the solution space. Consequently, the increased diversity among individuals enhances the algorithm's ability to adapt to different facets of the optimization landscape, potentially leading to more robust and effective solutions. It is obvious that transfer learning effectively leverages the feature extraction capabilities acquired on CIFAR-10 for application to CIFAR-100. Furthermore, this approach not only saves significant time but also enhances accuracy. 

During the fine-tuning process, various data augmentation techniques were employed, including Smoothing\cite{simonoff1996}, Mixup\cite{zhang2018mixup}, and AutoAugment (AA)\cite{cubuk2019autoaugment}. Smoothing involves attenuating the true labels from 1.0 to a smaller value (typically close to 1.0). This is a method to smooth the labels, which helps mitigate overfitting and enhances generalization performance. The main parameter is used to reduce the value of the true labels, for example, decreasing from 1.0 to 0.9. This value can be adjusted based on the specific task to balance the smoothness of the model and the fitting of the training data. Mixup: is a data augmentation technique that generates new training samples by linearly combining the inputs and labels of two different samples. This encourages the model to be more robust during learning. AutoAugment is an augmentation strategy that automatically discovers the best data augmentation policies for a given dataset. It optimizes augmentation policies to improve model robustness and generalization. AutoAugment typically involves a set of sub-policies, each containing a specific data augmentation operation and its corresponding probability. Key parameters include the number of sub-policies, the number of operations in each sub-policy, the intensity of each operation, etc. These parameters can be adjusted by searching on the training set or using default values.

Another direct approach to initializing the population for the DE  algorithm is to use the last-performing model obtained during a single fine-tuning process, which spans a duration of the last population size epochs, as the initial population. 
\section{Results}
\subsection{Hyperparameter effect}
Based on the experimental results, it was found that hyperparameters continue to have a significant impact on the performance of the DE algorithm. The experiments revealed that when the crossover rate (CR) and mutation factor (F) are relatively low, there is an improvement in the optimization of Spiking Neural Networks (SNNs) by DE. This finding contrasts with previous observations where DE was applied to optimize Artificial Neural Networks (ANNs), suggesting that DE requires smaller values of F and CR to effectively optimize SNNs.

\textbf{Influence of CR and F:} Experiments as shown \ref{tab:Change Hyperparameter} have demonstrated that lower values of CR and F tend to improve CADE's effectiveness in optimizing Spiking Neural Networks (SNNs). A lower CR means that a smaller portion of the trial individual's components will be inherited by the target individual during the crossover operation. A lower F value results in a more conservative amplification of the difference between parent individuals when generating the trial individual.

\begin{table}[ht]
\centering
\begin{tabular}{@{}cccccc@{}} % six columns with centered content
\toprule
F-init & CR-init & Strategy & t & Improvement & Accuracy \\ 
\midrule
2e-3   & 2e-3   & 2   & 1   & 0.05   & 77.81  \\
1e-5   & 1e-5   & 2   & 1   & 0.41   & 78.17  \\
1e-6   & 1e-6   & 2   & 1   & 0.19   & 77.95  \\
1e-7   & 1e-7   & 2   & 1   & 0.18   & 77.94  \\
1e-5   & 1e-5   & 2   & 2   & 0.42   & 78.18  \\
1e-6   & 1e-6   & 2   & 2   & 0.12   & 77.88  \\
1e-7   & 1e-7   & 2   & 2   & 0.12   & 77.88  \\
1e-5   & 1e-5   & 3   & 5   & \color[HTML]{003399} \textbf{0.52}   & 78.28  \\
1e-5   & 1e-5   & 3   & 5   & 0.33   & 78.09  \\
2e-3   & 2e-8   & 2   & 5   & 0.16   & 77.92  \\
1e-5   & 1e-5   & 2   & 10  & 0.41   & 78.17  \\

\bottomrule \\
\end{tabular}
\caption{Hyperparameters}
\label{tab:Change Hyperparameter}
\end{table}
Our experimentation with hyperparameter tuning in CADE demonstrates the algorithm's sensitivity and responsiveness to parameter adjustments. The ability to enhance performance through fine-tuning underscores the importance of hyperparameter optimization in achieving the full potential of evolutionary algorithms like CADE when applied to intricate neural network models such as SEW.

\subsection{Comparison of different initial population }

In our study, the initial population for the CADE framework was generated using three distinct strategies, each designed to optimize the diversity and robustness of the models. The first strategy involved fine-tuning models by adjusting different time ratios, allowing us to explore a range of computational efforts and model complexities. The second strategy segmented the CIFAR-100 dataset into five equal parts, with each segment used to train a unique model. This approach aimed to develop distinct capabilities in each model by exposing them to different data subsets. The third strategy employed a sophisticated transfer learning technique, where models were first trained on the CIFAR-10 dataset to harness broad generalizable features and then fine-tuned on CIFAR-100 to adapt these models to more specific tasks. This method effectively combines the benefits of generalization with specialized adaptation, enhancing the evolutionary potential of the population. These strategies and their impacts on optimization outcomes are comprehensively documented in Table III of our research findings.

The table \ref{tab:Comparison of different initial population} above illustrates the optimization outcomes of various initial populations in the CADE framework. It presents the performance metrics both before and after CADE's optimization process. Notably, it is observed that initial populations derived from pretrained models exhibit the most substantial enhancement in optimization. This comparison underscores the impact of different initial populations on the efficacy of the CADE optimization approach.
\begin{table}[ht]
\centering
\begin{tabular}{@{}c@{\hspace{2pt}}c@{\hspace{2pt}}c@{\hspace{8pt}}ccc@{}} % six columns with centered content
\toprule
  Initialization & Different  & Parts of CIFAR-100  & Pretrain Model   \\ 
   population&  Time Ratio & Datasets   &  from CIFAR-10  &   \\ 
\midrule
Acc Improvement   &  0.6  &0.01    & 0.65    \\
(From$\rightarrow$To) \%   &  76.24$\rightarrow$76.84  & 74.57$\rightarrow$74.58  & 77.37$\rightarrow$78.02    \\
\bottomrule \\
\end{tabular}
\caption{Different initial population}
\label{tab:Comparison of different initial population}
\end{table}
The initial population should ideally encompass a diverse range of solutions. This diversity ensures a broader exploration of the solution space, increasing the likelihood of finding globally optimal solutions rather than being trapped in local optima. In the context of machine learning and particularly with CADE, using pre-trained models as the initial population can be highly advantageous. Pretrained models are already trained on vast datasets and have learned general features that can provide a significant head start in the optimization process.

\subsection{Comparison with CADE, SADE and SHADE}
The notable performance improvement observed with CADE in optimizing the Spike-Element-Wise (SEW) model can be attributed to several factors.
% \begin{table}[ht]
% \centering
% \begin{tabular}{@{}cccccc@{}} % six columns with centered content
% \toprule
%   & DE & SADE  & SHADE & CADE \\ 
% \midrule
% Improvement   & 0.63   & 0.56   & 0.11   & {\color[HTML]{003399}\textbf{0.65}}     \\

% \bottomrule \\
% \end{tabular}
\begin{table}[ht]
\centering
\begin{tabular}{@{}ccccc@{}} % six columns with centered content
\toprule
  &  SADE  & SHADE & CADE \\ 
\midrule
Improvement   & 0.56   & 0.11   & {\color[HTML]{003399}\textbf{0.65}}     \\

\bottomrule \\
\end{tabular}
\caption{Comparision of DE variation}
\label{tab:Comparison with CADE, SADE and SHADE}
\end{table}
First of all, the introduction of the Cosine Adaptive Mechanism in CADE allows for a more nuanced adjustment of the F and CR. The cosine-based variation provides a smoother transition during the optimization process, preventing abrupt changes that might disrupt the delicate balance required for optimizing the SEW model.

In our comparative study, we analyzed the performance of two adaptive differential evolution algorithms, SHADE (Success-History Adaptation Differential Evolution) and SADE (Self-Adaptive Differential Evolution), based on the same initial population generated through our outlined strategies. The effectiveness of each algorithm was assessed by measuring the improvement in accuracy over the initial population. This approach allowed us to directly compare how each algorithm adapts and optimizes under identical starting conditions. By quantifying the accuracy improvements, we were able to evaluate the relative efficacy of SHADE and SADE in refining the models’ performance, providing a clear benchmark for evolutionary success within our CADE framework. This comparison is crucial in understanding which adaptation strategies yield the most gains in practical applications.

The comparative analysis may suggest that the self-adaptive mechanisms in SADE and adaptation of SHADE may not be as well-suited for the specific challenges posed by the optimization of SNNs. CADE, tailored with the cosine-based adaptive mechanism, proves to be more effective in navigating the complex optimization landscape associated with SEW.
In summary, the success of CADE in optimizing the SEW model can be attributed to its adaptability, smooth parameter variation, and its alignment with the unique characteristics of spiking neural networks. These qualities make CADE particularly well-suited for the challenges posed by the optimization of SNNs compared to SADE and SHADE.

\subsection{Robustness Experiment}
In the experimental phase, this paper not only compared the efficacy of CADE, SADE, and SHADE in optimizing the SEW model but also conducted robustness experiments to assess the performance of the CADE-optimized model against the original pretrained model.
\textbf{CADE-Optimized Model vs. Pretrained Model:} (the pretrained model is derived from the model obtained during the final popsize epochs of a fine-tuning process.) From the table~\ref{CADE-Optimized Model vs. Pretrained Model (Error)}, the CADE-optimized SEW model demonstrated robustness across varying corruption levels, maintaining competitive accuracy under different degrees of noise and perturbations. Despite the challenging conditions introduced by CIFAR-100-C, the CADE-optimized model exhibited the retention of adaptive features acquired during the optimization process, contributing to its resilience. In specific corruption types, the CADE-optimized model outperformed the pretrained model, suggesting that the optimization process enhanced the model's ability to handle certain types of distortions and corruptions.

\begin{table}[h]
\setlength{\tabcolsep}{4pt}
\renewcommand{\arraystretch}{1}
\footnotesize\centering
\begin{tabular}{l|ll|ll}

& \multicolumn{2}{c}{\textbf{Error}} &\multicolumn{2}{c}{\textbf{mCE}} \\
&{Finetune} & {CADE} & {Finetune} & {CADE} \\
& {SGD}&  &&{SGD}\\
 \hline
Gaussian Noise & 80.98\% &{\color[HTML]{003399} \textbf{80.94\%}} & 100\%  &{\color[HTML]{003399} \textbf{99.94\%}}   \\

Shot Noise & {\color[HTML]{003399} \textbf{75.43\%}} & 75.47\% & {\color[HTML]{003399} \textbf{100\%}} & 100.05\% \\
Impulse Noise & 80.50\% & {\color[HTML]{003399} \textbf{80.39\%}} & 100\% &  {\color[HTML]{003399} \textbf{99.86\%}} \\

Defocus Blur & 71.19\% & {\color[HTML]{003399} \textbf{71.16\%}}& 100\% & {\color[HTML]{003399} \textbf{99.95\%}} \\

Glass Blur &  91.60\% &{\color[HTML]{003399} \textbf{91.59\%}} & 100\% & {\color[HTML]{003399} \textbf{99.98\%}} \\
Motion Blur & 80.10\% & {\color[HTML]{003399} \textbf{80.01\%}} & 100\% & {\color[HTML]{003399} \textbf{99.88\%}}  \\
Zoom Blur & {\color[HTML]{003399} \textbf{80.70\%}} & 80.66\% &100\% & {\color[HTML]{003399} \textbf{99.94\%}}  \\
Snow & 62.59\% & {\color[HTML]{003399}62.45\%} &  100\% & {\color[HTML]{003399} 99.77\%} \\
Frost & 66.32\% & {\color[HTML]{003399} \textbf{66.16\%}} & 100\% & {\color[HTML]{003399} \textbf{99.76\%}} \\
Fog & {\color[HTML]{003399} \textbf{71.50\%}} & 71.60\% & {\color[HTML]{003399} \textbf{100\%}} & 100.13\% \\
Brightness & 45.71\% & {\color[HTML]{003399} \textbf{45.64\%}}& 100\% & {\color[HTML]{003399} \textbf{99.83\%}}   \\
Contrast & 80.96\% & {\color[HTML]{003399} \textbf{80.95\%}} & 100 \% & {\color[HTML]{003399} \textbf{99.98\%}}\\
Elastic Transform & {\color[HTML]{003399} \textbf{73.78\%}} & 73.82\% & {\color[HTML]{003399} \textbf{100\%}} & 100.05\%  \\
Pixelate &  67.30\% & 67.18\%& 100\% & {\color[HTML]{003399} \textbf{99.81\%}} \\
Jpeg Compression & 66.09\% &{\color[HTML]{003399} \textbf{66.11\%}} & {\color[HTML]{003399} \textbf{100\%}} & 100.03\%\\
 \hline
Average & 72.98\% &{\color[HTML]{003399} \textbf{72.94\%}} &100\% & {\color[HTML]{003399} \textbf{99.93\%}}\\ 
\end{tabular}
\caption{\textbf{Performance (error and mCE) on Cifar-100-C.} } 
\label{CADE-Optimized Model vs. Pretrained Model (Error)}

\end{table}
\textbf{mCE} - mean Corruption Error (mCE). To assess robustness, the experiments employ the mean Corruption Error (mCE) as a performance measure, which quantifies the error rate of the current model compared to the base model for various types of corruption.
% \[
% CE_{\text{Current\_Model}} = \frac{\sum_{s=1}^{5} E_{\text{Current\_Model},s,c}}{\sum_{s=1}^{5} E_{\text{Base\_Models},s,c}}
% \]

% In this equation, C represents the type of corruption, and S indicates the sensitivity level associated with that specific type of corruption. 

For both the CIFAR-100-C datasets, each corruption type comprises five distinct sensitivity levels. It's worth noting that while the base model employed in \cite{hendrycks19} was pretrain model finetuned by stochastic gradient descent algorithm, in our experiments, this paper use pretrain model as the base model. From the results, it is shown that the CADE algorithm can improve the robustness of SEW on CIFAR-100-C.

\subsection{Different CR and F update methods}
The experiment also involved contrasting the results of the trigonometric approaches with scenarios where the learning rate (LR) was fixed at its maximum and minimum values. This comparison was crucial to assess the effectiveness of dynamic versus static parameter settings.

\begin{table}[ht]
\centering
\begin{tabularx}{0.48\textwidth}{@{}ccccccc@{}} % six columns with centered content
\toprule
   & F\_sin\_CR\_sin  & F\_sin\_CR\_cos  & F\_cos\_CR\_sin & \\ 
\midrule
Acc Improvement   &  0.7            &0.7            & 0.55           \\
(From$\rightarrow$To)\%   &  77.37$\rightarrow$78.07  & 77.37$\rightarrow$78.07  & 77.37$\rightarrow$77.92 \\
\midrule
 &F\_cos\_CR\_cos &Fixed\_1e-5 &Fixed\_1e-9   \\ 
 \midrule
Acc Improvement   &0.65&0.63  &0.53   \\
(From$\rightarrow$To)    &77.37$\rightarrow$78.02 &77.37$\rightarrow$78 &77.37$\rightarrow$77.9   \\
\bottomrule \\
\end{tabularx}
\caption{Different CR and F update method}
\label{tab:Different CR and F update method}
\end{table}

The outcomes of these experiments demonstrated that the sine and cosine functions' application to F and CR variations is impactful. Both these trigonometric methods yielded better results compared to keeping the parameters fixed. It underscores the importance of adaptive parameter control in enhancing the performance of evolutionary algorithms.

\section{Discussion}
 The study noted that in SNNs, the weights often exhibit extreme values in comparison to traditional DNNs. This distinctive characteristic of SNNs implies that the optimization process needs to be handled with greater care. Aggressive exploration, potentially facilitated by higher values of CR and F, could disrupt the delicate balance within the structure of SNNs. As a result, a more nuanced and cautious approach to optimization is required to preserve the intricate dynamics of SNNs and avoid perturbing their unique structural characteristics.

\begin{table}[ht]
\centering
\begin{tabular}{c|ccc} % six columns with centered content
\toprule
 &Max & Min & Most value distribution interval  \\ 
\midrule
SEW   & 195300   & -10.2669   & [-0.03,0.03]    \\
ResNet-50      & 50040   & -1.0489  & [-0.04,0.05]\\
\bottomrule \\
\end{tabular}
\caption{Weight Distribution}
\label{tab:my_label}
\end{table}

\section{Conclusion}
This paper proposed the CADE algorithm, integrating DE principles with SNN characteristics to enhance the performance and robustness of SNNs. The CADE algorithm dynamically adjusts hyperparameters F and CR using cosine functions, significantly improving the Spike-Element-Wise (SEW) model's optimization efficacy. Our experiments demonstrated that CADE not only outperforms traditional DE, SADE, and SHADE in terms of optimization but also showcases remarkable enhancements in model robustness across diverse conditions, particularly in handling data corruption scenarios like CIFAR-100-C.

Reflecting on the broader implications of our findings, it was observed that SNNs often exhibit extreme values in weights compared to traditional Deep Neural Networks (DNNs), highlighting the need for a nuanced optimization approach. The delicate balance within SNN structures requires cautious and precise optimization strategies, as aggressive exploration through high values of CR and differential weight factor F could potentially disrupt the intrinsic dynamics of SNNs. This insight reinforces the need for adaptive and thoughtful approaches in the optimization of neural networks, particularly those that mimic biological processes more closely than conventional models.

% In conclusion, this paper proposed the CADE algorithm, which bybridizes DE principles with SNN characteristics to enhance the performance and robustness of SNNs. It delved into the design and operation of the CADE algorithm, including how it adjusts hyperparameters F and CR using cosine functions. Furthermore, it emphasizes the significant performance improvements achieved by CADE in optimizing the Spike-Element-Wise (SEW) model. Notably, CADE outperforms traditional DE, SADE, and SHADE algorithms in terms of optimization efficacy. Furthermore, this final year project highlighted the critical impact of initial population and parameter choices on the success of evolutionary algorithms, with CADE notably excelling in population optimization. Also,
% CADE has demonstrated significant improvements in enhancing the robustness of the SEW (Spike-Element-Wise) model. Through CADE optimization, the SEW model exhibits better generalization across various tasks and datasets. This implies that the SEW model can adapt more effectively to different application scenarios. CADE further strengthens the robustness of the SEW model, enabling it to maintain stable performance under data corruption (CIFAR100-C). This is particularly crucial for handling noise and variations in real-world applications.

\end{document}